\documentclass[11pt, a4paper]{article}
\PassOptionsToPackage{breaklinks}{hyperref}
\usepackage[hyperref]{emnlp2019}
\usepackage{times}
\usepackage{latexsym}

\usepackage{url}

\usepackage{url}
\usepackage{booktabs} 
\usepackage{graphicx}

\usepackage{array}
\usepackage{textcomp}
\usepackage{setspace}

\newcommand{\dset}{JuICe}

\usepackage{tikz}
\newcommand{\bluecheck}{}%
\DeclareRobustCommand{\bluecheck}{%
  \tikz\fill[scale=0.4, color=blue]
  (0,.35) -- (.25,0) -- (1,.7) -- (.25,.15) -- cycle;%
}

\aclfinalcopy 

\title{\dset: A Large Scale Distantly Supervised Dataset\\ for Open Domain Context-based Code Generation}

\author{Rajas Agashe, Srinivasan Iyer and Luke Zettlemoyer \\
  ∗Paul G. Allen School of Computer Science and Engineering, Univ. of Washington, Seattle, WA \\
  {\tt \{rajas, sviyer, lsz\}@cs.washington.edu}  \\}

\date{}

\begin{document}
\maketitle
\begin{abstract}
    Interactive programming with interleaved code snippet cells and natural language markdown is recently gaining popularity in the form of Jupyter notebooks, which accelerate prototyping and collaboration. To study code generation conditioned on a long context history, we present \dset, a corpus of 1.5 million examples with a curated test set of 3.7K instances based on online programming assignments. Compared with existing contextual code generation datasets, {\dset} provides refined human-curated data, open-domain code, and an order of magnitude more training data. Using \dset, we train models for two tasks: (1) generation of the API call sequence in a code cell, and (2) full code cell generation, both conditioned on the NL-Code history up to a particular code cell. Experiments using current baseline code generation models show that both context and distant supervision aid in generation, and that the dataset is challenging for current systems.    
  
\end{abstract}

\begin{figure}[t]
\includegraphics[width=\linewidth]{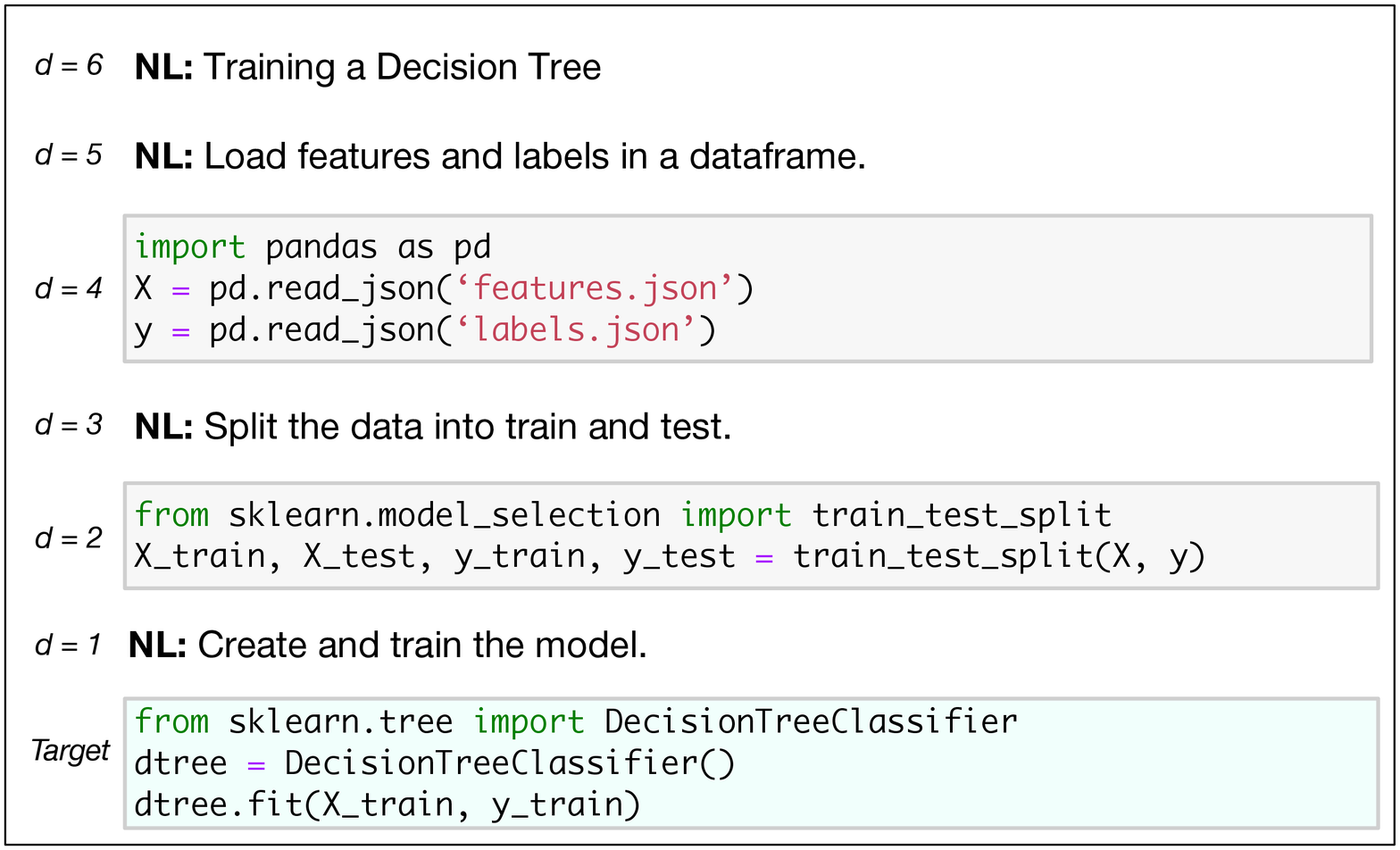}
\caption{This Python Jupyter notebook, comprising of interleaved NL markdown and code cells, loads data from a file and trains a decision tree classifier. We aim to generate the target code cell (blue) based on the previous NL markdown \textit{Create and train the model} and the prior NL and code history. To the left of each cell, \textit{d} represents its distance from the target cell.}
\label{fig:task_main}
\end{figure}

\section{Introduction}
Interactive computing (IC) is a software engidneering paradigm where programmers write source code scripts in an incremental fashion, one block at a time, taking decisions based on the output of execution of previously written blocks. 

An increasingly adopted platform for IC is the Jupyter notebook \cite{kluyver2016jupyter}, which additionally encourages the use of markdown in natural language (NL) between code snippets as a means of documentation (Figure \ref{fig:task_main}). Under the framework of IC, models that automatically generate future code blocks based on NL instructions, have the potential to provide significant assistance and speed up development. In this paper, we collect and release a new large-scale dataset based off of Jupyter notebooks named \dset{} (for Jupyter Interactive Computing). We also present experiments to demonstrate baseline performance for the task of incremental context-based code generation.

Figure \ref{fig:task_main} presents part of a Jupyter notebook for training a decision tree classifier using Python, which includes a title, followed by interleaved NL markdown and code cells. The markdown typically describes the goal of the code cells that follow and often is of high quality, since notebooks are frequently used for sharing and collaboration amongst teams. The first two code cells import packages and perform data loading, and following this, the user would typically inspect the data and proceed to train the model by first writing the markdown as \textit{Create and train the model}. Our goal is to train models that assist the user at this particular stage i.e. to generate the API calls \texttt{DecisionTreeClassifier()} and  \texttt{fit()} and  the source code contained in the blue cell in Figure \ref{fig:task_main} (which trains a Decision tree classifier), based on the previous markdown, and NL and code context. The closest NL utterance alone does not specify that a Decision Tree model must be trained, or which variables should be used as train and test data. Automatically making these decisions conditioned on prior history is one of the main challenges of this task.

Existing tasks for mapping NL to source code primarily use a single NL utterance \cite{zettlemoyer05,iyer-EtAl:2017:Long} to generate database queries (semantic parsing), single line python code \cite{yin2018mining,oda2015learning}, multi-line domain-specific code \cite{ling2016,rabinovich-stern-klein:2017:Long}, or sequences of API calls \cite{gu2016deep}. A recent task by \newcite{iyer2018} on the CONCODE dataset maps a single utterance to an entire method, conditioned on environment variables and methods. In contrast, we tackle the task of general purpose code generation in an interactive setting, using an entire sequence of prior NL and code blocks as context. More closely related to our task is the context dependent semantic parsing task on the ATIS dataset \cite{zettlemoyer2009learning,suhr2018learning} for mapping NL to database queries based on a prior history of NL and query pairs. One main difference is that while future queries can be built by modifying queries in previous utterances, code snippets in our task are almost always disjoint from previously generated code. 

We collect an open-domain large-scale dataset (\dset) of over 659K publicly available Jupyter notebooks from \url{github.com} to train models for our task. Each notebook contains an average of 39  NL/code cells and uses python packages from thousands of domains. Since obtaining source-code with NL annotations is expensive, most large NL-code datasets rely on noisy automatically scraped data from source code forums or repositories online \cite{yin2018mining,iyer2018}. However, this also results in excessively noisy test sets and thus, a persistent challenge has been the creation of high quality code generation test sets. For our dataset, we capitalize on the availability of nbgrader \cite{hamrick2016creating} notebooks, which are high-quality Jupyter notebooks that are manually created by instructors as programming assignments for classes. These notebooks contain accurate and detailed sequences of NL and code blocks, with blank code-cells intended to be filled in by students and later evaluated. We collect and release 13,905 nbgrader notebooks to evaluate models for code generation tasks, which, together with the training set, forms the first dataset for general purpose code generation in an interactive setting. 

We define two code generation related tasks using \dset: (1) generating API sequences, and (2) full code generation. The first task is more relaxed and aims to assist users by generating a sequence of all function calls \cite{gu2016deep} needed to achieve the goal of the cell. The second task measures the ability of models to learn to generate fully functioning complete code snippets. For both tasks, we investigate strong neural baselines and study performance improvements obtained by varying the size of prior NL and code context considered, as well as varying training data size. 

In summary, we introduce the task of code generation under the paradigm of IC. We collect and release a new large-scale dataset together with a manually curated test set based on nbgrader. Finally, we evaluate strong neural baselines on two code generation tasks, which achieve reasonable performance, with significant room for improvement.

\begin{table*}
\small
\centering
\resizebox{\linewidth}{!}{
\begin{tabular}{l|c|c|c|c}
\toprule
Dataset & Context Based & Open Domain & Large Scale & Curated Dev/Test \\
\midrule
\textbf{\dset{}} & \bluecheck & \bluecheck & \bluecheck & \bluecheck \\
\midrule
CONCODE \cite{iyer2018} & \bluecheck & \bluecheck  &  \bluecheck & \\
CoNaLa \cite{yin2018mining} & & \bluecheck &  \bluecheck & \bluecheck \\   
ATIS \cite{zettlemoyer2009learning} & \bluecheck &  &  & \bluecheck \\
SequentialQA \cite{iyyer2017search} & \bluecheck &  &  & \bluecheck \\
SCONE \cite{long2016simpler}& \bluecheck &  &  & \bluecheck \\          
\bottomrule
\end{tabular}}
\caption{Comparison of {\dset} with various recently released NL-code datasets along four dimensions. Most datasets that condition on historical natural language and code are small and domain specific, while large scale datasets only use human curated test sets for simple single line NL-code generation.}
\label{tab:comparison}
\end{table*}

\section{Related Work}

There is significant existing research on mapping single NL utterances directly to executable programs in the form of logical forms \cite{zettlemoyer05}, $\lambda$-DCS \cite{liang2011learning}, regular expressions \cite{kushman-barzilay:2013:NAACL-HLT,locascio-EtAl:2016:EMNLP2016}, database queries \cite{iyer-EtAl:2017:Long,zhong2017seq2sql} and general purpose programs \cite{balog2016deepcoder,allamanis2015bimodal}. 
\newcite{ling2016} generate Java and Python source code from NL for card games, conditioned on categorical card attributes. \newcite{iyer2018} generate Java class methods using a single NL instruction and conditioned on a list of class environment variables and methods. \newcite{yin2018mining} mine a large NL-code dataset from Stackoverflow to train models to map a NL programming question into a short example solution snippet. \newcite{Gu:2016:DAL:2950290.2950334} use neural models to map Java documentation strings to a sequence of API calls. In this work, we introduce a new task of mapping NL to source code under an interactive programming paradigm i.e., conditioned on a sequence of past NL and code cells representing past exploratory interactions.

Models for mapping NL to code have been evaluated on datasets containing templated code for card games (Hearthstone \& MTG; \citeauthor{ling2016}, \citeyear{ling2016}), or manually labeled per-line comments (DJANGO; \citeauthor{oda2015learning}, \citeyear{oda2015learning}). These datasets contain $\sim$20,000 programs with short textual descriptions possibly paired with categorical data and are highly domain specific with limited context. In this work, we collect the first large scale dataset containing over 1.5M source code cells paired with a sequence of NL and code cells that represents past exploratory interactions. Although various large scale datasets \cite{allamanis2013mining,allamanis2014mining,allamanis2016convolutional,iyer2018} to study code generation have been created from Github, their development and test set are randomly created from the same dataset since human curation is prohibitively expensive. Similarly, \newcite{yin2018mining} collect a large dataset from Stackoverflow.com (CoNaLa) for training, but only manage to curate a small portion ($\sim$ 2,900 examples) of single line NL and code snippets for evaluation. We take advantage of nbgrader assignment notebooks to create an inexpensive high-quality human-curated test set of 3,725 NL statements with interactive history.

Neural encoder-decoder models have proved effective in mapping NL to logical forms and also for directly producing general purpose programs. \newcite{ling2016} use a sequence-to-sequence model with attention and a copy mechanism \cite{gu-EtAl:2016:P16-1} to generate source code. Recent methods focus on constrained decoding mechanisms to generate syntactically correct output using a decoder that is either grammar-aware or has a dynamically-determined modular structure paralleling the structure of the abstract syntax tree of the code \cite{dong2016,rabinovich-stern-klein:2017:Long,krishnamurthy-dasigi-gardner:2017:EMNLP2017,yin-neubig:2017:Long,iyer2018}. \newcite{iyer2018} use a specialized context encoder that uses sub-word units \cite{sennrich-haddow-birch:2016:P16-12} for representing code tokens and a grammar-aware decoder that attends to both NL and context to produce source code parses. Although syntax aware models are more accurate, they are significantly slower than Seq2Seq to train on large datasets. We present strong Seq2seq baselines with sub-word units for our task, that take interactive context into account and scale to large datasets.

\begin{table}[h]
\small
  \begin{center}
  \resizebox{\linewidth}{!}{
    \begin{tabular}{l|c|c}
      \toprule 
      & Train & Dev/Test \\
      \midrule
      \# Examples & 1,518,049 & 1,744/1,981 \\

      Avg  Context Cells &        29.9  &         28.3 \\
        Avg NL Tokens       &         39.6 &         57.2 \\
        Avg Code Tokens     &         38.8 &         33.7 \\

      \# Unique NL Token  &   851,127      & 11,142 \\
      \# Unique Code Token  & 1,001,289     & 5,113 \\

      \midrule
      \% Use variables above & 45.3 & 58.6 \\
      \% Use methods above & 6.4 & 8.2 \\
      \% Contextual & 48.2 &  61.9 \\
      \% Multi-cell & 19.7 & 29.0 \\

      \bottomrule 
    \end{tabular}}
    \caption{Statistics for the {\dset} dataset. The dev/test sets are obtained from high-quality programming assignments. Expanded descriptions for the bottom four rows are: \% Examples using variables defined above, \% Examples using methods declared above, \% Examples using a variable or method from above (i.e. contextual), \% Examples where the variables/methods come from at least 2 different code cells. To accurately measure contextual reasoning, we only consider those variables/methods that are not mentioned in the NL above.}
    \label{tab:stats}
  \end{center}
\end{table}

\section{Dataset Collection}

We aim to study code generation in an interactive setting, conditioned on prior history of NL and code cells. Jupyter notebooks are a popular platform for exploratory programming that encourage users to write code one cell at a time and to make programming decisions by executing previous cells and examining the output. Since these notebooks are heavily used for sharing and collaboration, they are typically well documented with high quality NL. The intent of the NL (markdown) is typically to describe the functionality of the code cells that follow, as opposed to code comments, which primarily describe why code is written in a certain way. We collect a large scale training set to serve as distant supervision for our human-curated dev/test sets based on programming assignments and exercises.

\subsection{Training Set Creation}

To create {\dset} we first collect all publicly available Jupyter notebooks from \url{github.com} created before May 2019 and filter for notebooks having NL markdown in English and Python 2/3 as their kernel type. We observe that the presence of NL markdown is correlated with notebook quality and remove any notebooks that have more than three times the number of code cells as the number of NL cells, leaving us with $\sim$ 659K notebooks. 

Within these notebooks, each code cell is a potential training example, and we define the sequence of NL and code cells above it as its context. We only consider code cells that have NL markdown in the immediate previous cell (henceforth, target NL), as target cells. For our dataset, we include target cells that have syntactically valid Python and at most one method definition (for cells containing methods). Following \newcite{iyer2018}, we canonicalize strings in the target cell and truncate them to a maximum of 120 code tokens. This amounts to over 4 million (target cell, context) examples that can be used for training, which we downsample to 1.5 million by ensuring that each target cell and context is unique. Table \ref{tab:stats} presents dataset statistics for {\dset}.

\subsection{Human Curated Dev/Test Set}
\label{sec:devtest}

\begin{figure}[t]
\includegraphics[width=\linewidth]{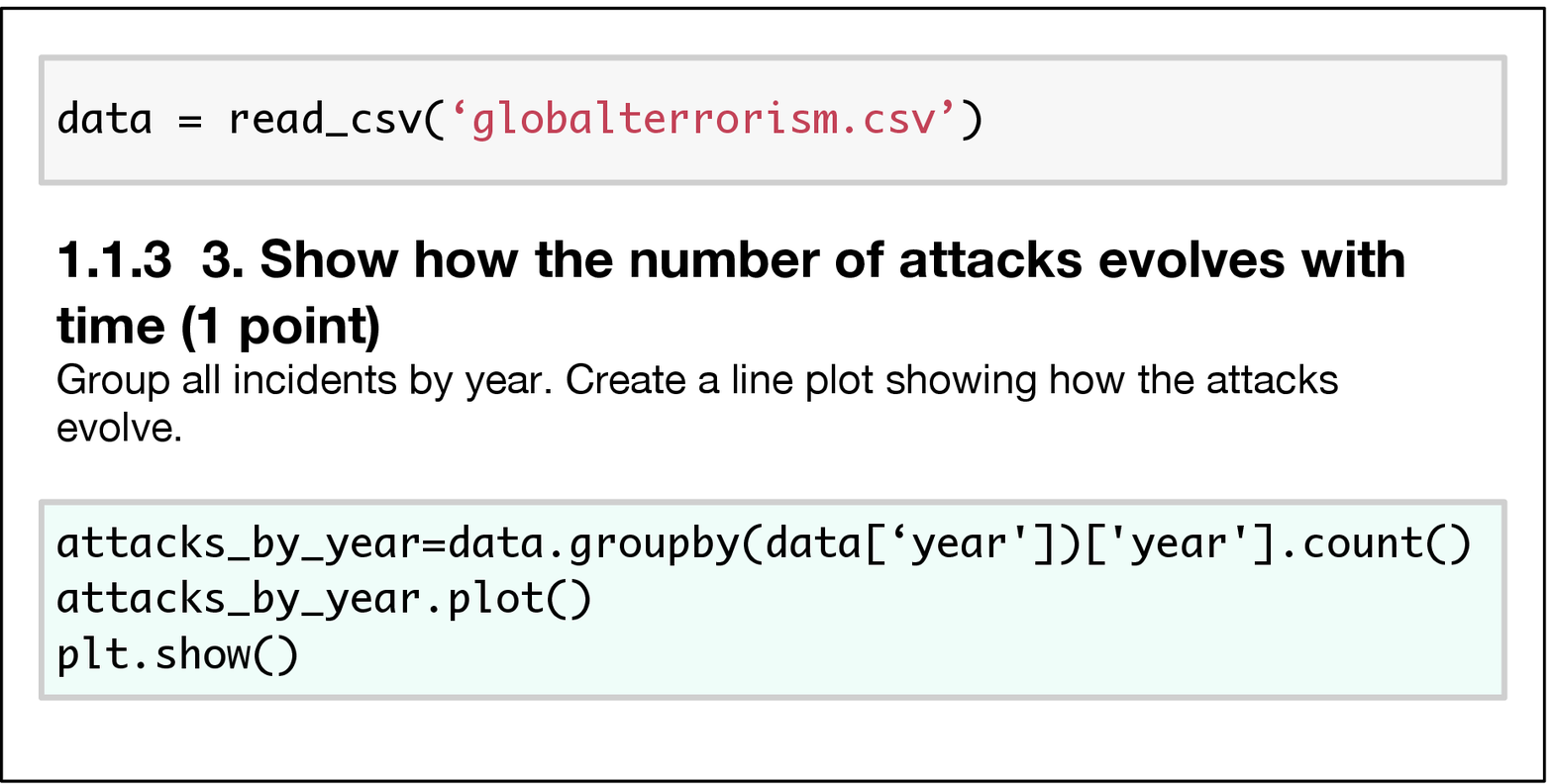}
\caption{This example nbgrader problem prepared by an instructor prompts the student to fill in the blue colored cell with code to perform operations on previously loaded data. We leverage such notebooks with embedded solutions, posted on Github, to create a high quality dev and test set for \dset.}
\label{fig:nbgrader_example}
\end{figure}

While existing large scale NL-code datasets have leveraged noisy online code repositories/forums to create training sets, obtaining a human curated test set has always been prohibitively expensive, and is typically done only for simple single line NL to code tasks. For our dataset, we take advantage of programming assignments written in Jupyter notebooks to build a high quality dev/test set. Specifically, we collect 1,510 code context examples from assignments created with nbgrader \cite{hamrick2016creating}, and mine an additional 2,215 examples from in-class exercise notebooks.

\textbf{Nbgrader Assignments}\space \space \space  Instructors use nbgrader to create and grade programming assignments within Jupyter notebooks. Figure \ref{fig:nbgrader_example} presents one such nbgrader assignment, where the student must wrangle some data that has been loaded earlier in the notebook to create a plot, and enter the solution in the cell in blue. Behind the scenes, nbgrader stores the following properties within each code cell's metadata: (1) the number of points its worth, (2) a unique code cell ID, (3) whether it's to be autograded or manually graded, (4) boilerplate code, (5) whether it is an instructor solution or a student submission. While students are expected to confidentially send their completed assignment to instructors, the solutions are often later posted on Github. 

Taking advantage of this, we search Github for nbgrader notebooks with embedded metadata and collect 13,905 such notebooks, each containing multiple prompts and target code cells, with multiple student submissions for the same assignment. Based on the problem IDs and point values, we group together all student submissions that answer the same instructor prompt. Each prompt written by an instructor has an average of 8.9 different student submissions. If the metadata of any solution notebook indicates a successful execution on all the test cases without generating errors in the output, we select that as one of the correct submissions. If there are multiple correct submissions we select one at random, and wrong solutions are discarded. If the instructor provided boilerplate within the target cell, it is extracted and placed into the context above. Thus, each prompt is paired with one solution, which forms our target cell.

\textbf{In-Class Exercises}~ While instructors use nbgrader for creating formal programming assignments, they often use vanilla Jupyter notebooks for creating informal in-class exercises. In this case, an instructor can release an exercise notebook intended for students to complete and a companion solution notebook to provide immediate feedback. The students are free to solve the exercise and upload the notebooks to their git repositories.

We use the following procedure to extract high quality solution target code cells from these notebooks. First, we collect notebooks containing the word \textit{solution} in the title, and pair it with an exercise notebook from the same repository if over 50\% of the cells are the same and \textit{solution} is not in the exercise notebook's title. We find 170K such notebook pairs. If corresponding code cells in the two notebooks differ, then the exercise cell contains the student solution and the other, the instructor solution. We consider these cells as potential target cells, and to verify the instructor solution, we confirm that the same notebook and solution exist in at least 2 other Github repositories.

We find a total of 1,510 nbgrader assignment problems and 2,215 in-class exercise problems and split them randomly into the dev and test sets. We perform the same checks on the target cell as we do on the training set and additionally convert any solutions written in Python 2 to Python 3. Table \ref{tab:stats} shows the dataset statistics. We release this, together with our training set, as the first large scale dataset to study code generation with prior NL and code history\footnote{https://github.com/rajasagashe/juice}. Table \ref{tab:comparison} compares {\dset} with various recently released code generation datasets on four different dimensions: interactive history, domain specificity, scale, and human annotation.   

\section{Data Statistics}
Our quantitative and qualitative analysis of examples in {\dset} show that the NL is diverse, the code broadly covers data science applications, and that context is important for generating the target cells.

\begin{table*}[h]
\small
  \begin{center}
    \begin{tabular}{l|c|p{11cm}}
      \toprule
      NL Type & \% & Example  \\
      \midrule
      High-level Declarative & 43 & Analyze the Average prices(open, high, low, close) for all stocks in the Month of May 
       \\

      Variables/Functions & 27 &  Convert all categorical variables in \texttt{bank\_features} into indicator variables\\

      Arguments Mentioned & 20 &  Use unigrams, bigrams, trigrams ... english stopwords \\

      Question & 18 &  Which is the cheapest video game ?\\
      
      Line-by-line & 16 & Create a histogram. Use default settings. Add the label...\\    

      Long References & 11 & 3d ) Transform reviews...using the tfidf vectorizer we created in part 2 . Save the transformed...\\    
      
      Input/Output & 9 & ...Given strings a and b, return a single string with a and b separated by a space, except swap the first 2 chars of each string. e.g. (\texttt{mix, pod} $\rightarrow$ \texttt{pox mid} and (\texttt{dog, dinner} $\rightarrow$ \texttt{dig, donner}) \\

      Equation & 7 & Solve the system $C_\epsilon \theta_\epsilon = b_\epsilon$ for $\theta_\epsilon$. Store the result in a variable, \texttt{theta\_eps}. \\

      \bottomrule
    \end{tabular}
    \caption{Qualitative analysis of NL from the dev set of {\dset}. A large percentage of the NL is abstract and describes the code at a high level. Furthermore, instructors use a variety of additional modalities to elucidate the prompt, including input-output examples and equations.}
    \label{tab:nl}
  \end{center}
\end{table*}

\paragraph{NL Analysis} Since assignment prompts tend to be more detailed in order to provide specific instructions for students, the average NL length in dev/test is $\sim$20 tokens longer than that in train (Table \ref{tab:stats}). Table \ref{tab:nl} presents statistics of 50 target NL cells classified into overlapping categories. A high percentage (43\%) of the NL is abstract language in the declarative or interrogative form, while 27\% explicitly state which variables/functions to use, and 16\% have a line-by-line correspondence between each line of language and code. For the latter two categories we believe that getting full exact match is possible since the procedure is clearly specified. The last three rows of Table \ref{tab:nl} present interesting NL phenomenon. 11\% refer to markdown far above in the same notebook, 9\% show examples of the desired inputs and outputs of the code to clarify challenging language, and 7\% use mathematical equations to formally specify the intended mathematical operations to be performed.

\begin{table}[h]
\small
  \begin{center}
    \begin{tabular}{l|c}
      \toprule
      Code Type & \%  \\
      \midrule
      Data Exploration & 25  \\
      Data Wrangling & 23 \\
      Machine Learning & 20 \\
      Miscellaneous & 16  \\
      Visualization &  13   \\
      Systems & 3 \\

      \bottomrule
    \end{tabular}
    \caption{{\dset} includes code for various real world applications, primarily for data science/machine learning.}
    \label{tab:codetype}
  \end{center}
\end{table}

\paragraph{Code Analysis} Table \ref{tab:codetype} presents statistics of 50 code cells sampled from the dev set. Majority of code cells are geared towards data science applications. The \textit{Data Exploration} category covers scenarios for computing statistics or queries to gain new insights about the data, and the \textit{Machine Learning} category covers training models, as well as tensor operations. 13\% of code cells are used for plotting figures and the \textit{Miscellaneous} category covers a broad range of utility functions.

\paragraph{Quality Evaluation} 
We estimate the quality of our dataset using a sample of 50 examples from our train and dev splits. A good quality example is one in which there is enough signal to generate the full target code cell using the NL and the context. We find that 68\% examples from train and 96\% examples from dev are solvable. Most of the noise in the training set arises from the target NL referring to another code cell (for example, the cell above the NL). Other examples require sophisticated background knowledge e.g. the mathematical processes behind music composition.

\paragraph{Contextual Reasoning} An analysis of 50 examples reveals the diverse types of contextual reasoning required to generate the target cell. Most examples require using variables or methods defined above (similar to statistics in Table \ref{tab:stats}). 39\% require understanding the creation and usage of data structures in the context. For example, in row 2 of Table \ref{tab:contextreasoning}, knowing that \texttt{unemployment} stores the rates for every quarter in the array, suggests the use of the \texttt{diff} function which finds the difference between every consecutive value. In other cases, knowing the column names of tables is necessary and these are often specified as positional arguments while loading the data. Being able to condition on loaded table columns and values as in \newcite{krishnamurthy-dasigi-gardner:2017:EMNLP2017} using stored in-notebook variable outputs is an interesting area for future work. 25\% examples require reusing idioms specific to the notebook to tailor the code to the programmer's style or to leverage efficient primitives that the user has defined. For example, the implementation for \texttt{countEvenDigits} (row 3) can reuse the efficient list comprehension defined in a similar method above. Lastly, 23\% of examples require multi-cell reasoning where the model must incorporate elements from at least 3 cells. In row 4 of Table \ref{tab:contextreasoning}, the \texttt{GridSearch} object created in one cell needs to be used to select the best hyperparameter values, and retrain the model using the same procedure found in another code cell. 

Finally, using rule-based metrics, we find that 61\% examples are contextual i.e. use a token from the context (from Table \ref{tab:stats}), whereas 86\% examples required contextual reasoning that includes reasoning from data and idioms as well.

\begin{table*}[h!]
\small
  \begin{center}
  

    \begin{tabular}{m{3cm}|c|c}
    
      \toprule
      Type & \% & Example \\
      
      \midrule
       Incorporates \textbf{Variables} or \textbf{Functions} declared or used above. & 48 &  
       \begin{minipage}{.7\textwidth}
      \includegraphics[width=\linewidth]{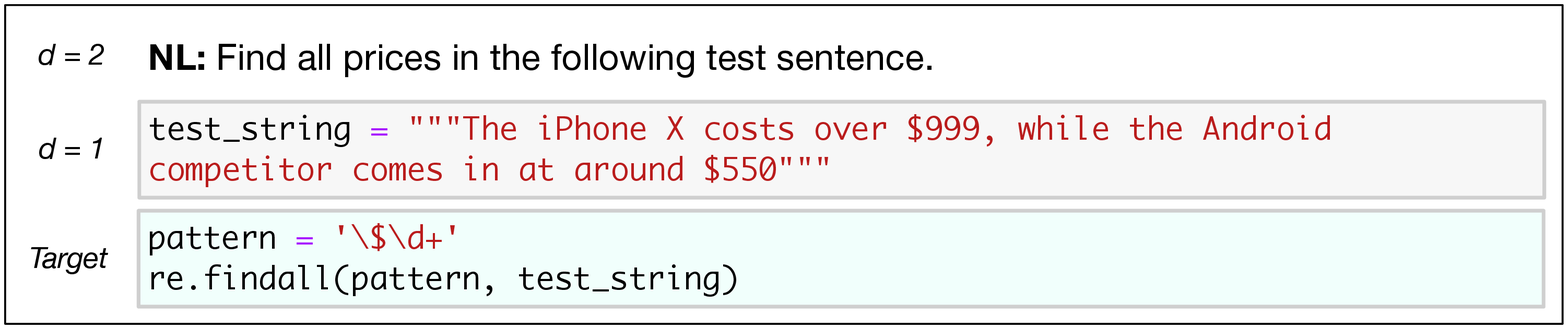}
      \end{minipage} \\

      \midrule
      Requires understanding the properties of the \textbf{data} and its underlying structure to perform logic.  & 39 &  
      \begin{minipage}{.7\textwidth}
      \includegraphics[width=\linewidth]{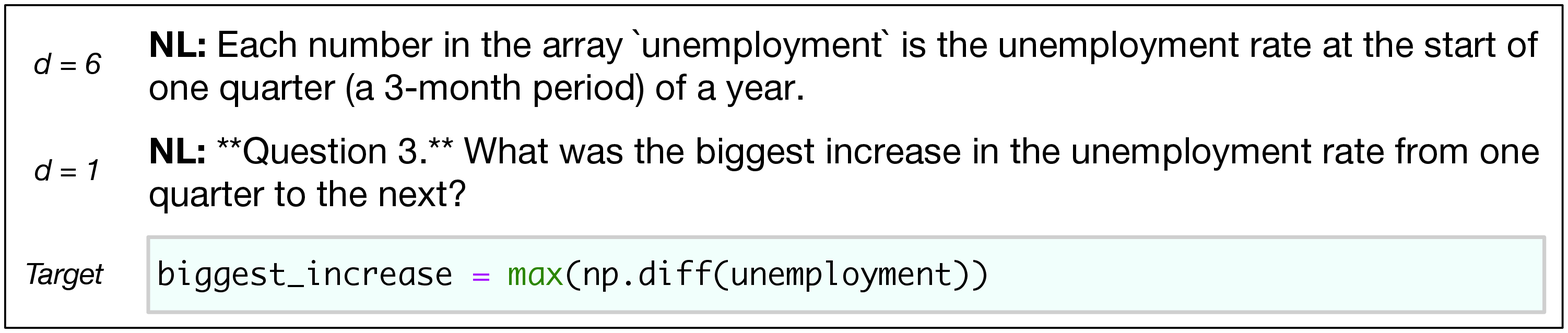}
      \end{minipage} \\

      \midrule
      Incorporates coding patterns or \textbf{idioms} specific to the notebooks.  & 25 &
      \begin{minipage}{.7\textwidth}
      \includegraphics[width=\linewidth]{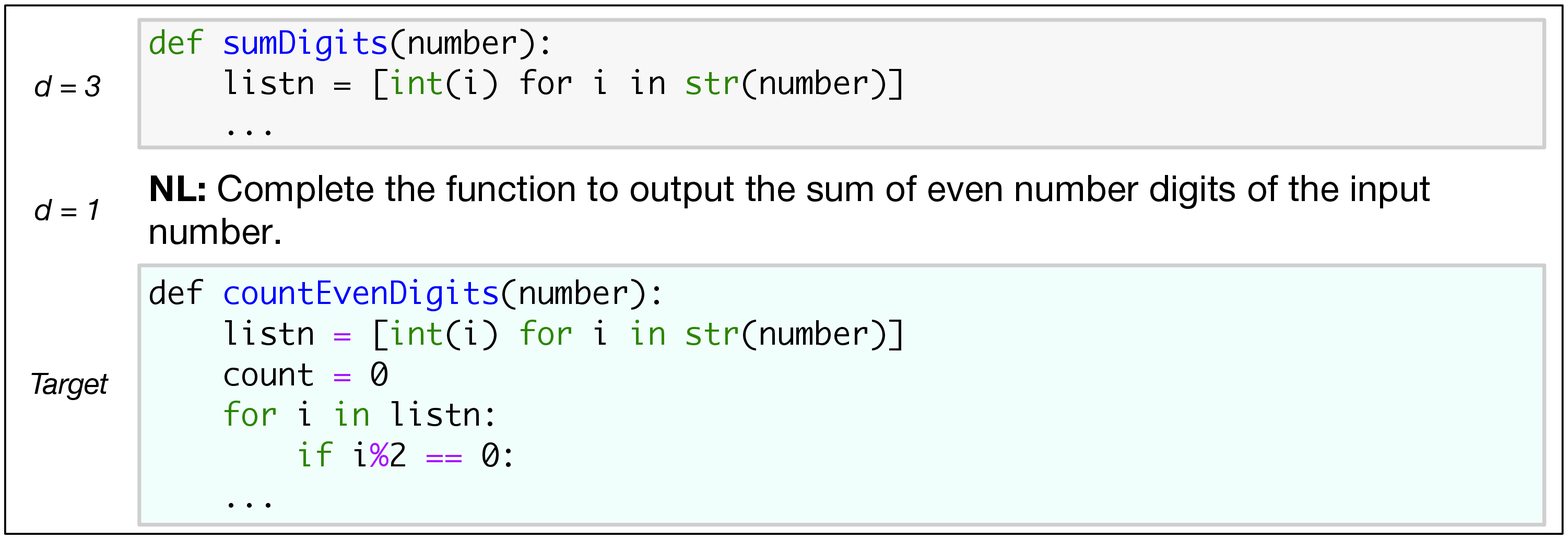}
      \end{minipage}
      \\
      
      \midrule
      \textbf{Multi-cell reasoning} involves reasoning over at least 3 context cells. & 23 &
      \begin{minipage}{.7\textwidth}
      \includegraphics[width=\linewidth]{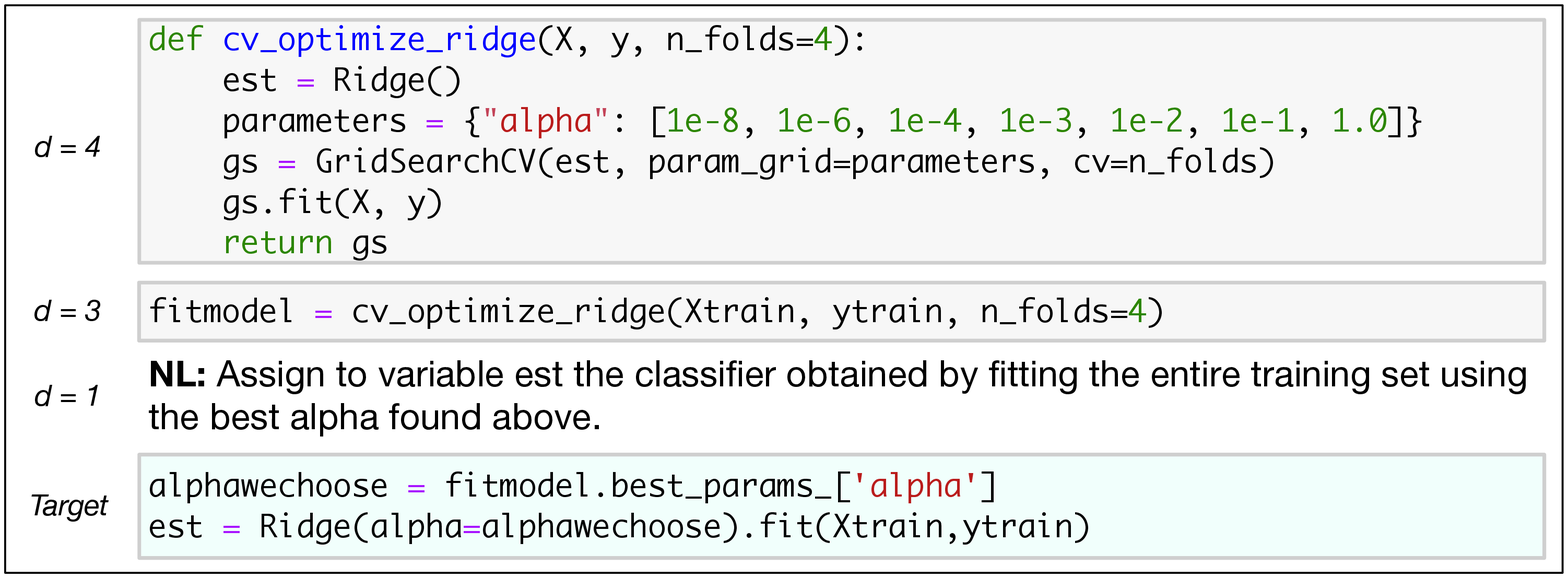}
      \end{minipage}
      \\
      \bottomrule
    \end{tabular}
    
    \caption{Types of contextual reasoning required for target cell generation. Overall, 86\% of examples require contextual reasoning.}
    \label{tab:contextreasoning}
  \end{center}
\end{table*}

\section{Problem formulation}
We investigate two tasks for target cell generation in notebooks, conditioned on the context cells up to the target cell: (1) The Full code generation task, which involves generating the entire code snippet in the target cell. (2) The API sequence task, which involves generating only the sequence of constructor and method calls within the target cell, without having to generate variables and arguments. This task is more relaxed and can itself be very helpful to developers under an interactive setting. We do not aim to generate keyword functions such as \textit{range} as these are very commonly used. Figure \ref{fig:task_main} presents an example for our code generation task, where our goal is to learn models that will map the NL \textit{Create and train the model}, together with the context cells above, into either the the full code snippet (in blue) that reuses previously defined variables \texttt{X\_train} and \texttt{y\_train}, or the API sequence, \textit{DecisionTreeClassifier fit}.

Formally, let $a^{(i)}$ denote the API sequence, and $s^{(i)}$ denote the code token sequence of the  target cell in the $i$th training example. Further, if $c_d^{(i)}$ represents the context cell at a distance $d, d > 0$, above the target cell of example $i$, our tasks are to generate $a^{(i)}$ and  $s^{(i)}$, conditioned on the context cells. Note that $c_d^{(i)}$ can be a sequence of NL words or code tokens. In models that we present in later sections, we limit $d$ to a maximum of $K$ cells.

\section{Baselines}

We train and evaluate a selection of neural (using fairseq: \newcite{ott2019fairseq}) and non-neural baselines that are known to perform well for code generation tasks, particularly on large scale datasets. 

\subsection{Retrieval}
This baseline presents the programmer with an existing code cell from elsewhere in the training corpus as a possible prediction for the target code cell, located using only the NL markdown immediately before the target cell i.e. $c_{1}$. We evaluate two variants: (1) Ret-Train uses the code cell underneath the NL markdown cell from the entire training set, that is most similar to $c_{1}$ , (2) Ret-Context uses the code cell underneath the most similar NL cell amongst all $c_{d}$ (context above). The similarity between two NL cells is computed by representing both NL sequences as vectors using tf-idf representations for their dimensions and measuring the cosine distance between these vectors. We convert the retrieved code cell to an API sequence for the API sequence task.

\subsection{LSTM with Attention}
This baseline is a neural encoder-decoder model where the encoder computes contextualized representations of input sequence embeddings using an n-layer BiLSTM, and an LSTM-based decoder produces a sequence of code (API) tokens, while attending to the encoder representations at every time step $t$. We make use of the general global attention mechanism by \newcite{luong2015effective}. To use this model for our task, we concatenate all $K$ context cells, $c_{1\dots K}$, using the type of each cell as separator symbols i.e. \texttt{CODE} or \texttt{MARKDOWN}. The output sequence is supervised to be either an API sequence or a sequence of code tokens.

\subsection{Transformer}
This baseline utilizes the Transformer \cite{vaswani2017attention} architecture where the encoder and decoder consist of multiple layers of multi-headed self-attention and position-wise feed forward layers. The inputs and outputs are the same as the LSTM model described above.

\begin{table}[t]
\centering
\resizebox{\linewidth}{!}{
\begin{tabular}{l|c|c|c|c|c|c}
\toprule
 & & & \multicolumn{2}{c|}{Dev} & \multicolumn{2}{c}{Test} \\

 & & & \multicolumn{2}{c|}{Full Code} & \multicolumn{2}{c}{Full Code} \\
Models & K & N & Bleu & EM & Bleu & EM \\

\midrule
Ret-Train & - & 100,000 & 5.52 & 0.82 &  4.76 & 0.48 \\
Ret-Context & $\vert c \vert$ & - & 3.42  & 0.12 & 3.24  & 0.00 \\ 
\midrule
Transformer & 1 & 100,000 & 3.57 & 0.00 & 3.21  & 0.00 \\
Transformer & 3 & 100,000 & 10.90 & 0.27 & 11.08  & 0.38 \\
LSTM   & 1 & 100,000 & 7.35 & 0.05 & 7.92 & 0.14  \\

LSTM & 3 & 100,000 & 15.88 & 1.42 & 17.03 & 1.33 \\

\textbf{LSTM} & \textbf{3} & \textbf{1,518,049} & \textbf{21.66} & \textbf{5.57}  & \textbf{20.92} & \textbf{5.71} \\

\bottomrule
\end{tabular}}
\caption{Exact Match and BLEU score for the full code generation task on both the dev and test sets of {\dset} for all baselines. All models benefit with additional code context ($K=3$), which permits conditioning on variables and methods defined previously. Training on additional data further pushes up performance.}
\label{tab:results}
\end{table}

\begin{table}[t]
\centering
\resizebox{\linewidth}{!}{
\begin{tabular}{l|c|c|c|c|c|c}
\toprule
 & & & \multicolumn{2}{c|}{Dev} & \multicolumn{2}{c}{Test} \\

 & & & \multicolumn{2}{c|}{API} & \multicolumn{2}{c}{API} \\
Models & K & N & Precision & Recall & Precision & Recall \\

\midrule
LSTM   & 1 & 100,000 & 34.15 & 29.19 & 30.38 & 26.98  \\
LSTM & 3 & 100,000 &  37.09 &  31.48 & 37.44 &  33.23 \\

\textbf{LSTM} & \textbf{3} & \textbf{1,518,049} & \textbf{51.34} & \textbf{44.83}  & \textbf{52.60} & \textbf{46.46} \\
\bottomrule
\end{tabular}}
\caption{Precision and Recall for the API sequence task on the dev/test sets of {\dset} for all baselines}
\label{tab:resultsapi}
\end{table}

\section{Experiments}

We run code generation experiments to evaluate our baselines on {\dset} and to study the effects of context size $K$ and dataset size $N$.

To study the effect of context, we consider two configurations for our neural approaches, $K=1$ which consists of only the NL markdown above the target cell, and $K=3$. For both cases, we truncate the contents of each context cell to 75 tokens, except for the target NL cell, which remains as is. We consider $K=\vert c \vert$ for the retrieval baseline as it is allowed to access all context cells above. We train these configurations for the Seq2Seq models on a subset of 100K examples from the training set, and use the best settings to train on the full training set.

\subsection{Hyperparameters}

For Seq2Seq models, we use an embedding size of 1024 for the input and output tokens. We consolidate the NL and code vocabularies and apply BPE with 10,000 merges. Both encoder and decoder LSTMs use 2 layers with a hidden size of 1024. We use dropout $p = 0.5$ in between LSTM layers and over the decoder output. The model is trained for 40 epochs using gradient based methods using the Adam optimizer \cite{kingma2014adam} with a learning rate of 0.001, and we use beam search based decoding with a beam size of 5 during inference.  The Transformer model uses 6 layers and 4 attention heads. The dimension of the model is 512 and the dimension of the feed forward network is 1024. We use a learning rate of 0.0005 with 2,000 warmup steps, train for 200 epochs, and pick the model at the epoch with highest validation performance.  

\subsection{Metrics}

Similar to prior code generation tasks, we use Exact Match (EM) accuracy and corpus-level BLEU score \cite{papineni2002bleu} as performance metrics for full code generation. While EM is a strict metric measuring the ability of models to generate fully functioning code, BLEU serves as a measure of partial credit for models that can provide users with a partially correct snippet.

For the API sequence task, we find BLEU and EM inadequately measure performance since the average API sequence length for our target code is 3.8. We therefore treat the predicted API calls as a set of values and compute precision/recall metrics against the gold set of API calls.

\section{Results and Discussion}

\begin{figure}
    \centering
\resizebox{.75\linewidth}{!}{\includegraphics{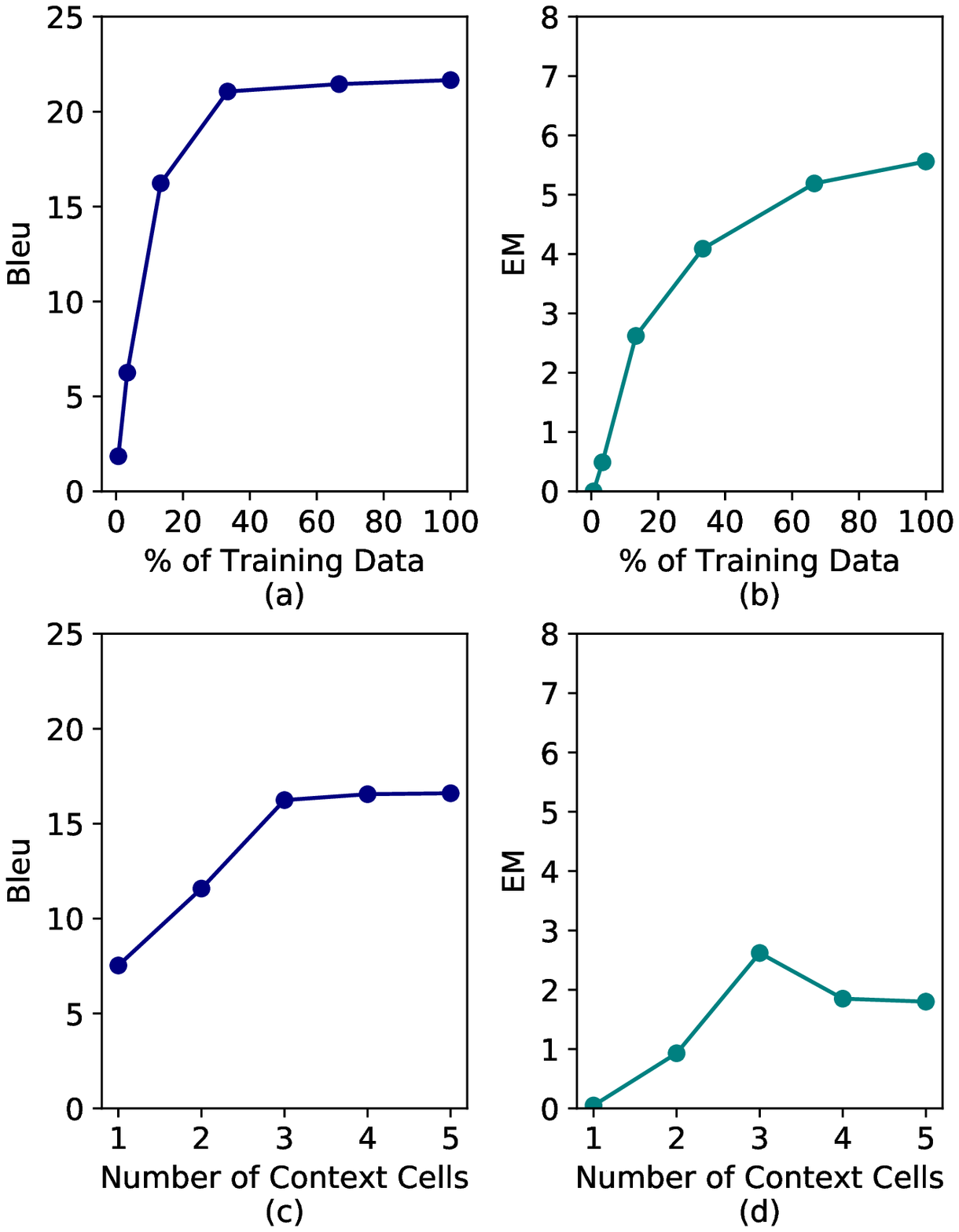}}   
    \label{fig:my_label}
    \caption{(a) and (b) show the effects of training data on LSTM(3-ctx) performance. Feeding the model more \dset{} consistently improves EM accuracy. (c) and (d) show the effects of context cells on performance of LSTM trained with 200K examples. Performance improves for up to 3 cells, after which reasoning over long contexts becomes challenging for current methods.}
\label{fig:datactxgraph}
\end{figure}

We present EM and BLEU scores for the code generation task in Table \ref{tab:results}, and precision/recall for the API sequence task in Table \ref{tab:resultsapi}, on both the dev and test sets of {\dset}  for all our baselines. 

All LSTM models outperform both retrieval baselines by more than 2\% BLEU score. We find that increasing the amount of training data and the context length helps all models for both tasks. Our best model was trained on 1.5 million examples with a context length of 3 (Table \ref{tab:results}). The improvement with context is most likely owing to the interactive nature of the notebook, where variables (methods) defined in previous code cells are operated upon (called) in future cells. However, note that the retrieval model which is given access to the full context performs significantly worse, suggesting that the code above is different than the target code and the model needs to reason over which pieces of it to incorporate.

Figure \ref{fig:datactxgraph} illustrates the effects of varying training data size and context length on our LSTM model. We find that increasing the amount of training data helps significantly up to 200K examples, after which, performance begins to plateau. Furthermore, the model improves with more context up to 3 cells and then struggles to incorporate additional cells as the average input sequence length exceeds 156 tokens. 

Using student notebook solutions for a subset of the test set, we estimate a human performance of 60\% BLEU and 23\% EM (can be much higher with identifier canonicalization) for the code generation task, and 84\% precision and 85\% recall for the API sequence task.

\subsection{Error Analysis}

We conduct a qualitative error analysis using 50 erroneous predictions of our top performing model - \textbf{LSTM K=3, N=1,518,049} on dev set examples (Table \ref{tab:error}). We find that for 39\% of cases, the model misunderstands the core intent of the NL and generates totally incorrect code. In 17\% cases, it misses some positional arguments in the code. For example, \textit{Use unigrams, bigrams, and trigrams. Use English stop words} should result in a \texttt{Vectorizer} object with arguments: \texttt{stop\_words="english", ngram\_range=(1,3)}. 26\% of cases represent an inadequate understanding of the context and 10\% require more cells than the 3 presented to the model. Retrieving relevant context cells far above the target is an interesting area for future work. 26\% cases are partially correct i.e. they contain some correct code lines but miss some details. For example, for the NL \textit{Create a histogram of all people's ages. Use the default settings. Add the label Age on the x-axis and Count on the y-axis}, the model generates the code for plotting the histogram but misses labeling the axes. For 15\% of cases the model generates semantically equivalent code but is different at the surface level. In the future, executing the code to compare outputs or verifying that the nbgrader tests pass would alleviate this issue but requires the model to generate valid strings (currently strings are canonicalized).

\begin{table}[t!]
\small
  \begin{center}
    
    \begin{tabular}{l|l}
      \toprule
      Error Category  & \% \\
      \midrule
        Challenging NL Reasoning & 39 \\
        Arguments Missed & 17 \\
        Contextual Reasoning & 26 \\
        Needs Longer Context & 10 \\ 
        Partially Correct & 26 \\
        Semantically Equivalent & 15 \\

      \bottomrule
    \end{tabular}
    \caption{Qualitative error analysis on 50 incorrectly generated code cells from our dev set for our best performing baseline. The first category represents cases where the model output was totally incorrect. Over 30\% of errors require deeper contextual reasoning.}
    \label{tab:error}

  \end{center}
  
\end{table}

\section{Conclusion}
In this paper, we introduced the task of code generation under the paradigm of interactive computing, conditioned on a context of interleaved code snippet cells and NL markdown. To train models for this task, we collected a new large scale open-domain dataset ({\dset}) from publicly available Jupyter notebooks, consisting of target code cells paired with sequences of NL and code context. Furthermore, we gathered a high quality evaluation set using nbgrader  and in-class programming assignment notebooks with solutions to reliably test code generation models. We evaluated a variety of baseline  models  for  two  context  dependent  code generation tasks, API sequence generation and full code generation. Experiments showed that performance improves on using an increased amount of code context and training data, with significant room for improvement.

\section*{Acknowledgements}
The research was supported in part by the ARO (ARO-W911NF-16-1-0121) and the NSF (IIS-1252835, IIS-1562364).
The authors thank the anonymous reviewers for their helpful comments.

\bibliography{emnlp2019}
\bibliographystyle{acl_natbib}

\end{document}